\pdfoutput=1

\documentclass[11pt]{article}

\usepackage[]{ACL2023}

\usepackage{times}
\usepackage{latexsym}

\usepackage[T1]{fontenc}

\usepackage[utf8]{inputenc}

\usepackage{microtype}

\usepackage{inconsolata}

\usepackage{multirow}
\usepackage{graphicx}
\usepackage{colortbl} 
\usepackage{tikz}     
\usepackage{xcolor}
\usepackage{enumitem}
\usepackage{booktabs}
\usepackage{adjustbox}
\usepackage{subcaption} 
\title{\textit{CoinMath}: Harnessing the Power of Coding Instruction for Math LLMs}

\author{
Chengwei Wei\textsuperscript{$\diamondsuit$}, 
Bin Wang\textsuperscript{$\diamondsuit$}, 
Jung-jae Kim \textsuperscript{$\diamondsuit$}, 
Guimei Liu \textsuperscript{$\diamondsuit$}, 
Nancy F. Chen\textsuperscript{$\diamondsuit,\dag$}
\\
\textsuperscript{$\diamondsuit$}Institute for Infocomm Research (I$^2$R), A*STAR, Singapore\\
\textsuperscript{$\dag$}Centre for Frontier AI Research (CFAR), A*STAR, Singapore\\
\texttt{wei\_chengwei@i2r.a-star.edu.sg} \\
}

\begin{document}
\maketitle
\begin{abstract}
    Large Language Models (LLMs) have shown strong performance in solving mathematical problems, with code-based solutions proving particularly effective. However, the best practice to leverage coding instruction data to enhance mathematical reasoning remains underexplored. This study investigates \textbf{three key questions}: (1) How do different coding styles of mathematical code-based rationales impact LLMs' learning performance? (2) Can general-domain coding instructions improve performance? (3) How does integrating textual rationales with code-based ones during training enhance mathematical reasoning abilities? \textbf{Our findings} reveal that code-based rationales with concise comments, descriptive naming, and hardcoded solutions are beneficial, while improvements from general-domain coding instructions and textual rationales are relatively minor. Based on these insights, we propose CoinMath, a learning strategy designed to enhance mathematical reasoning by diversifying the coding styles of code-based rationales. CoinMath generates a variety of code-based rationales incorporating concise comments, descriptive naming conventions, and hardcoded solutions. Experimental results demonstrate that CoinMath significantly outperforms its baseline model, MAmmoTH, one of the SOTA math LLMs.\footnote{Our model, code, and datasets are open-sourced at \url{https://github.com/amao0o0/CoinMath}}
\end{abstract}

\section{Introduction}

    Large Language Models (LLMs) have been extensively applied to solving mathematical problems \cite{lewkowycz2022solving, azerbayev2023llemma, luo2023wizardmath, ahn2024large, shao2024deepseekmath}. Recently, the approach of using code-based solutions has achieved significant success, as the calculations are handled by a program interpreter, while the reasoning steps are effectively represented in code format. Prior studies on prompting, such as Program-of-Thoughts (PoT)  \cite{chen2022program} and Program-Aided Language models (PAL) \cite{gao2023pal}, have demonstrated that generating solutions in a code format significantly enhances LLM performance on mathematical tasks compared to Chain-of-Thoughts (CoT) which uses a text format \cite{wei2022chain}. Furthermore, LLMs trained on mathematical instruction data with code-based rationales, reasoning steps presented in executable code, have shown strong performance \cite{wang2023mathcoder, yue2023mammoth, zhang2024infinitymath, toshniwal2024openmathinstruct}.

    Given the demonstrated effectiveness of mathematical instruction data with code-based rationales, there is growing interest in synthesizing additional mathematical problems to expand mathematical coding instruction datasets \cite{zhang2024infinitymath, toshniwal2024openmathinstruct}, aiming to enhance the performance further. Despite this progress, the question of how such coding instruction data can be best leveraged to maximize its impact remains underexplored. Recently, \citet{zhang2024unveiling} showed that model instruction-tuned on text or a mix of text and code outperforms models trained solely on code in mathematical reasoning. However, the conclusions are drawn from evaluating the models by generating solutions in a text format instead of a code format. \citet{bi2024program} found that code-based rationales of medium complexity yield the best reasoning performance. 
    
    Taking this further, our study delves deeper into the effective utilization of existing coding instruction data to enhance mathematical reasoning. Specifically, we first conduct a comprehensive study on coding instruction to address three critical questions: 1) How do different coding styles (e.g., having a detailed comment or not) of mathematical code-based rationales impact LLMs' learning performance in mathematical reasoning? 2) Can coding instructions from general domains, beyond math, provide meaningful benefits under PoT reasoning? and 3) How does integrating textual rationales with code-based ones during instruction tuning enhance mathematical reasoning abilities? 
    
    Our findings can be summarized as follows:
    \begin{itemize}[topsep=1pt, partopsep=1pt, itemsep=1pt]
      \item Mathematical code-based rationales that include concise comments and descriptive naming conventions are the most effective. Hardcoded rationales offer straightforward solutions that facilitate the model's learning.
      \item General domain coding instructions provide limited performance improvement compared to those from the mathematical domain.
      \item Adding textual rationales of math questions slightly enhances the performance of general-purpose models. In contrast, textual rationales do not benefit code-specialized models.
    \end{itemize}
    
    As the impact of coding instructions from general domains and textual rationales is minimal, while models exhibit varying behaviors toward code-based rationales with different coding styles, we propose \textbf{CoinMath}, short for \textbf{Co}ding \textbf{In}struction for \textbf{Math}, a learning strategy to effectively enhance LLMs’ mathematical reasoning capabilities. CoinMath diversifies the coding styles of code-based rationales by incorporating advantageous coding attributes, including concise comments, descriptive naming, and hardcoded solutions. Experimental results demonstrate that CoinMath significantly outperforms the previous SOTA Math LLMs.

    Our major contributions are as follows: 
    \begin{itemize}[topsep=1pt, partopsep=1pt, itemsep=1pt] 
        \item We present a systematic study that investigates what makes diverse coding instructions effective, how they enhance the mathematical reasoning of LLMs, and why they work, uncovering the key factors behind their impact. 
        \item Based on the above findings, we propose CoinMath, a learning strategy that improves LLMs' mathematical reasoning by incorporating code-based rationales with concise comments, descriptive naming conventions, and hardcoded solutions. Experimental results show that it outperforms the SOTA model by an average improvement of 5.9\% in accuracy.
        \item We release the CoinMath model, curated datasets, and training and evaluation pipelines, enabling reproducibility and advancing research in mathematical reasoning for LLMs.
    \end{itemize}

\section{Related Work}

    \textbf{Solving Math with Code.}
    Chain-of-Thought (CoT) prompting \cite{nye2021show, wei2022chain}, which guides LLMs to generate intermediate steps in textual rationales, has proven effective on reasoning tasks. Expanding on this foundation, Program-of-Thoughts (PoT) and Program-Aided Language models (PAL) \cite{chen2022program, gao2023pal} have used code for reasoning. These methods prompt LLMs to generate reasoning steps as code and assign computations to external program interpreters, which has achieved significant success, particularly in mathematical reasoning tasks that involve complex calculations and logical operations. Subsequently, recent works \cite{toshniwal2024openmathinstruct, wang2023mathcoder, yue2023mammoth, gou2024tora} have constructed mathematical instruction datasets with code-based rationales and fine-tuned LLMs on them, further improving model performance on solving math problems.

    \noindent \textbf{Effect of Coding Instruction on Mathematical Reasoning.}
    Several studies have explored coding instruction’s impact on LLMs' mathematical reasoning abilities. \citet{wang2023far} and \citet{ma2024at} indicated that coding instruction only enhances task-specific reasoning capabilities, which suggests non-math-specific data have minimal or even negative effects on mathematical reasoning. \citet{zhang2024unveiling} found that models trained on pure textual instructions or a combination of textual and coding instructions perform better in mathematical reasoning than those trained solely on coding instructions. However, the evaluations of these studies focused on CoT reasoning and generating textual solutions rather than code-based ones, which may limit the potential benefits of coding instructions. By contrast, \citet{bi2024program} evaluated models in PoT reasoning and revealed that mathematical code-based rationales with medium complexity lead to the most significant improvements. However, they did not comprehensively examine the effects of different coding styles, code domains, or the interplay between textual and code-based rationales in enhancing mathematical reasoning.

\section{Study on Coding Instruction} \label{sec:coding instruction study}
    This study investigates the effective utilization of coding instructions to enhance the mathematical reasoning abilities of LLMs. It focuses on three key questions. 1) Coding Style: Identifying the effect of different coding styles of mathematical code-based rationales on LLMs' mathematical reasoning learning. 2) Code Domain: Exploring whether coding instructions from general (non-mathematical) domains can enhance performance in solving mathematical problems under PoT reasoning. 3) Integration with Textual Rationales: Evaluating the impact of combining mathematical textual rationales with code-based ones during training to enhance mathematical performance.

    To explore the above questions, we collect instruction tuning datasets for three types: \textbf{Math-Text}, \textbf{Math-Code} and \textbf{General Code}. The Math-Text dataset, derived from MathInstruct~\cite{yue2023mammoth}, contains 26k math questions annotated with textual rationales. Similarly, Math-Code is another subset of MathInstruct, consisting of the same 26k math questions as Math-Text but annotated with code-based rationales. Finally, General Code comprises 21k Python coding instructions for non-math-specific tasks, curated from Code Alpaca \cite{codealpaca} and code generation tasks\footnote{\url{https://huggingface.co/datasets/iamtarun/python_code_instructions_18k_alpaca}}. Detailed statistics for these datasets can be found in Appendix Table \ref{table:it data}.
    
    \begin{figure*}[t]
        \centering
        \includegraphics[width=0.93\textwidth]{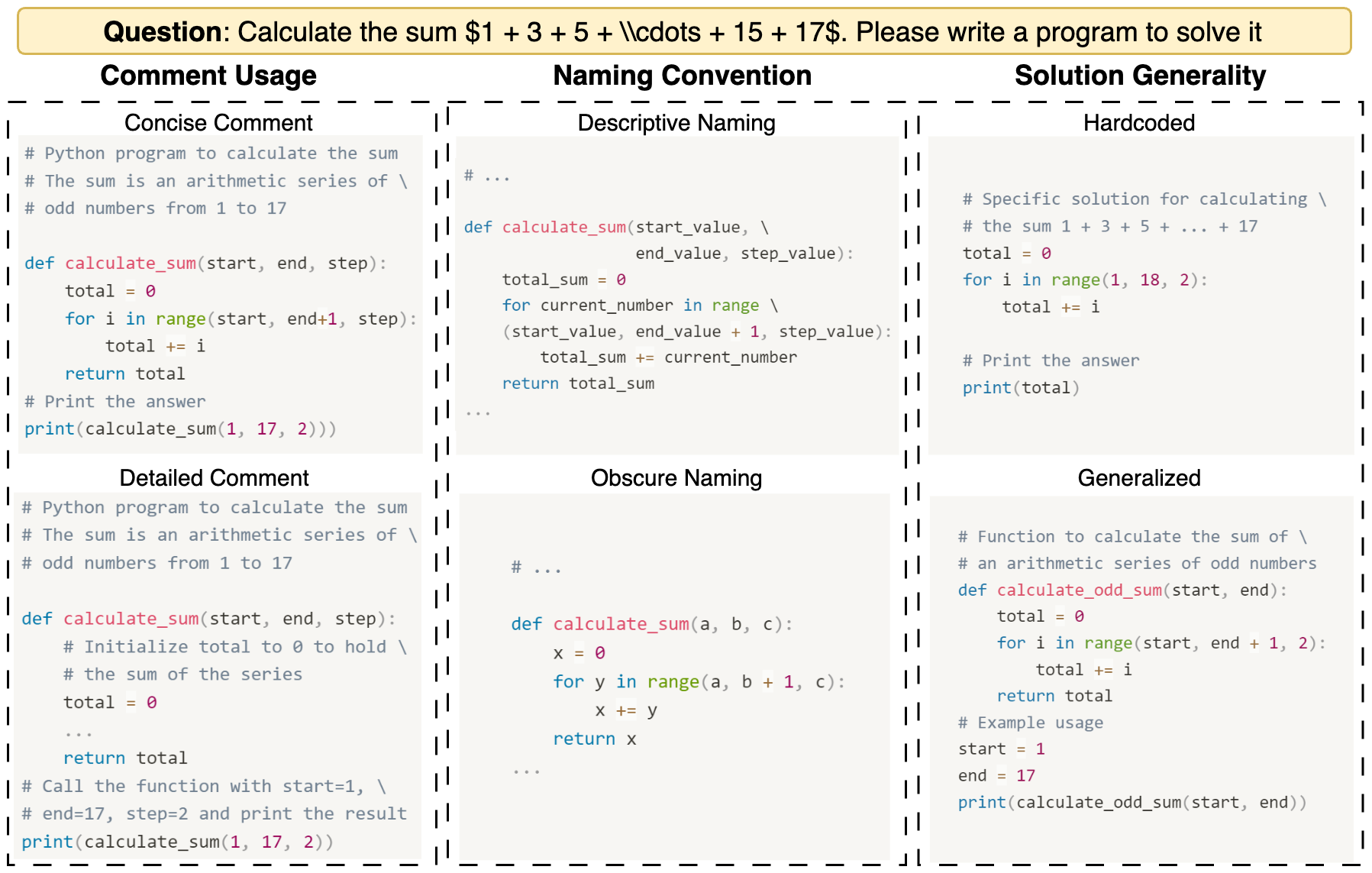}
        \caption{Exemplary code-based rationales in various coding styles (excluding No Comment in Code Notation). Certain lines are folded or omitted for improved visualization.}
        \label{fig:code style}
    \end{figure*}
    
    For evaluation datasets, we use four math Q\&A datasets representing distinct mathematical perspectives: (1) \textbf{Arithmetic}: A dataset of arithmetic questions from MathBench \cite{liu2024mathbench}. (2) \textbf{SVAMP} \cite{patel2021nlp}: A primary-school-level algebraic word problem dataset. (3) \textbf{GSM} \cite{cobbe2021training}: A middle-school-level algebraic word problem dataset. (4) \textbf{MATH} \cite{hendrycksmath2021}: A challenging dataset that extends beyond the high school level and covers diverse topics, including algebra, precalculus, and number theory. Examples from each dataset are provided in Appendix Figure~\ref{fig:example_eva_set}. While Arithmetic focuses solely on basic arithmetic, the difficulty of mathematical problems increases progressively from SVAMP to GSM, and finally to MATH. By default, we use PoT reasoning during evaluation unless otherwise specified as CoT reasoning.

    Both general-purpose and code-specialized models are evaluated. The general-purpose models include Llama-3.1-8B \cite{dubey2024llama} and Gemma-2B \cite{team2024gemma}, while the code-specialized models include CodeLlama-Python-7B \cite{roziere2023code} and CodeGemma \cite{team2024codegemma}. The instruction tuning details are provided in Appendix~\ref{sec:appendix_instruction_tuning_details}.

    \subsection{Coding Style} \label{subsec: study on coding styles}
    What coding styles in code-based rationales are most effective for model learning? To address this question, we examine three key attributes of coding styles that may influence a model's mathematical reasoning capabilities: Comment Usage, Naming Convention, and Solution Generality. Previous studies have discussed the role of variable naming and comment usage in prompting design~\cite{chen2022program, gao2023pal}. 
    In contrast, we focus on the influence of the instruction tuning phase with three attributes including a newly introduced one called, Solution Generality. 
    Figure~\ref{fig:code style} illustrates examples of different coding styles across these attributes. To generate rationales with diverse coding styles, we employ GPT-4o \cite{openai2024hello} to transform the original code-based rationales in the Math-Code dataset into variations reflecting diverse coding styles. Detailed instructions for generating these rationales are provided in Appendix~\ref{sec:appendix_prompt}.
    
    We focus on evaluating models in a zero-shot setting, as instruction-tuned models trained with datasets featuring different coding styles typically require few-shot examples that align with those styles, but a few-shot setting with different examples for different models may lead to biased evaluation results. Zero-shot evaluation allows us to isolate the effects of the instruction tuning data as the sole variable.

    \begin{table}[b]
        \centering
        \begin{adjustbox}{width=0.44\textwidth,center}
        \begin{tabular}{c | c | c c c c | c}
        \hline
        Model & Comment & A. & S. & G. & M. & Avg. \\
        \hline
        \multirow{3}{*}{\shortstack{Llama-3.1-\\8B}} 
         & No & 61.7 & 30.4 & 32.4 & 11.5 & 34.0 \\
         & Concise & 83.3 & 80.3 & 73.5 & 36.4 & \textbf{68.4} \\ 
         & Detailed & 80.7 & 83.4 & 73.6 & 35.5 & 68.3 \\
        \hline
        \multirow{3}{*}{\shortstack{CodeLlama-\\Python-7B}} 
        & No & 82.7 & 64.4 & 52.2 & 27.3 & 56.7 \\
        & Concise & 83.0 & 68.7 & 59.7 & 28.1 & \textbf{59.9} \\ 
        & Detailed & 83.3 & 69.6 & 55.4 & 27.2 & 58.9 \\
        \hline
        \multirow{3}{*}{\shortstack{Gemma-2-\\2B}} 
        & No & 73.0 & 65.3 & 53.6 & 25.8 & 54.4 \\
        & Concise & 77.3 & 67.6 & 54.7 & 26.6 & \textbf{56.6} \\ 
        & Detailed & 79.0 & 67.2 & 54.0 & 24.4 & 56.2 \\
        \hline
        \multirow{3}{*}{\shortstack{CodeGemma-\\2B}} 
        & No & 79.0 & 53.1 & 40.9 & 21.4 & 48.6 \\
        & Concise & 80.0 & 52.0 & 41.0 & 21.7 & 48.7 \\ 
        & Detailed & 81.0 & 55.2 & 39.7 & 21.4 & \textbf{49.3} \\
        \hline
        \end{tabular}
        \end{adjustbox}
        \caption{Results with different Comment Usages. 
        The best average performance for each model is highlighted in bold. 
        A., S., G., and M. refer to the Arithmetic, SVAMP, GSM, and MATH datasets, respectively.} \label{table:code style 1}
        \end{table}
    
    \noindent\textbf{Comment Usage.}
    We analyze the role of comment density in code-based rationales. Comments explain the overall logic of the rationales and the reasoning behind each step. We create three types of code-based rationales with varying comment usages: No Comment, Concise Comment, and Detailed Comment. In No Comment, the code stands alone, requiring the model to learn the code's functionality and mathematical logic solely from the code. Concise Comment includes essential annotations that clarify key steps in math problem-solving. Detailed Comment rationales offer comprehensive annotations, providing line-by-line explanations of the code's functionality and the underlying mathematical logic. By examining these three types of comment usage, we can assess how the presence and density of comments in instruction data influence the model's mathematical reasoning capabilities.

    The experimental results are summarized in Table \ref{table:code style 1}. Compared with No Comment, both Concise Comment and Detailed Comment styles exhibit good overall performance, with Concise Comment slightly outperforming Detailed Comment. Additionally, the performance improvement of Concise Comment and Detailed Comment over No Comment is larger for general-purpose models than for the code-specialized models. This suggests that while both general-purpose and code-specific models benefit from comments, code-specialized models inherently have a strong ability to understand code, making comments less critical for them. In contrast, comments are relatively crucial for general-purpose models to better grasp the mathematical reasoning underlying the code.

    \noindent\textbf{Naming Convention.}
    We identify naming conventions that best support models' mathematical reasoning learning. We design two types of naming conventions, namely, Descriptive Naming and Obscure Naming, while keeping other aspects of the code, such as comments and logic, unchanged. Descriptive Naming uses clear and meaningful names for all variables, whereas obscure naming employs short, non-descriptive names (e.g., single letters or random abbreviations).

    \begin{table}[t]
        \centering
        \begin{adjustbox}{width=0.45\textwidth,center}
        \begin{tabular}{c | c | c c c c|c}
        \hline
        Model & Naming & A. & S. & G. & M. & Avg. \\
        \hline
        \multirow{2}{*}{\shortstack{Llama-\\3.1-8B}} 
        & Descriptive & 82.3 & 81.8 & 72.9 & 35.6 & \textbf{68.2} \\
        & Obscure & 78.0 & 80.1 & 72.3 & 35.7 & 66.5 \\ 
        \hline
        \multirow{2}{*}{\shortstack{CodeLlama-\\Python-7B}} 
        & Descriptive & {80.3} & 69.8 & {58.3} & 28.0 & \textbf{59.1} \\
        & Obscure & 79.0 & 66.1 & 55.5 & 28.0 & 57.2 \\ 
        \hline
        \multirow{2}{*}{\shortstack{Gemma-\\2-2B}} 
        & Descriptive & {74.7} & {65.3} & {57.8} & {26.5} & \textbf{56.1} \\
        & Obscure & 73.7 & 62.3 & 55.0 & 25.9 & 54.2 \\ 
        \hline
        \multirow{2}{*}{\shortstack{CodeGemma-\\2B}}
        & Descriptive & {74.0} & {57.3} & {40.9} & {21.6} & \textbf{48.4} \\
        & Obscure & 72.0 & 54.8 & 39.7 & 21.4 & 47.0 \\ 
        \hline
        \end{tabular}
        \end{adjustbox}
        \caption{Results with different Naming Conventions.} \label{table:code style 2}
    \end{table}

    As shown in Table \ref{table:code style 2}, the results indicate that Descriptive Naming generally outperforms Obscure Naming. These findings are consistent with previous studies on naming conventions in prompting \cite{chen2022program, gao2023pal}. This further supports the idea that Descriptive Naming provides semantic context, helping models associate variable and function names with their intended purposes, which in turn enhances their understanding of mathematical problems.
        
    \noindent\textbf{Solution Generality.}
    We explore the benefits of generalized versus hardcoded solutions. In a generalized solution, a reusable function is written to capture the logic underlying the math problem, allowing the code to handle similar mathematical problems by accepting different inputs. 
    Such reusable functions help models acquire generalizable knowledge of mathematical problems, thereby improving their ability to generalize. In contrast, hardcoded solutions are tailored to specific problem instances, making them straightforward and easier for the model to learn.

    As shown in Table \ref{table:code style 3}, hardcoded solutions consistently outperform generalized solutions, suggesting that models benefit more from the explicit specificity of hardcoded logic, which simplifies learning. However, these results do not imply that hardcoded solutions are always better for mathematical instruction tuning. For example, a retrieval-augmented generation (RAG) model \cite{gao2023retrieval} may perform better with generalized solutions if it retrieves relevant functions effectively. We do not explore this area, as our experiments focus on a zero-shot setting for coding styles.

    \begin{table}[t]
        \centering
        \begin{adjustbox}{width=0.45\textwidth,center}
        \begin{tabular}{c | c | c c c c | c}
        \hline
        Model & Generality & A. & S. & G. & M. & Avg. \\
        \hline
        \multirow{2}{*}{\shortstack{Llama-3.1-\\8B}} 
        & Hardcoded & {83.3} & {78.9} & {73.5} & {34.8} & \textbf{67.6} \\
        & Generalized & 78.0 & 75.1 & 72.7 & 29.0 & 63.7 \\ 
        \hline
        \multirow{2}{*}{\shortstack{CodeLlama-\\Python-7B}} 
        & Hardcoded & {82.0} & {67.9} & {58.5} & {29.3} & \textbf{59.4} \\
        & Generalized & 71.3 & 67.4 & 54.8 & 21.7 & 53.9 \\
        \hline
        \multirow{2}{*}{\shortstack{Gemma-2-\\2B}} 
        & Hardcoded & {78.3} & {62.5} & {55.6} & {26.4} & \textbf{55.7} \\
        & Generalized & 68.7 & 60.3 & 51.4 & 18.9 & 49.8 \\ 
        \hline
        \multirow{2}{*}{\shortstack{CodeGemma-\\2B}} 
        & Hardcoded & {76.0} & {52.8} & {40.3} & {21.4} & \textbf{47.6} \\
        & Generalized & 68.7 & 49.0 & 39.2 & 16.9 & 43.5 \\
        \hline
        \end{tabular}
        \end{adjustbox}
        \caption{Results with different Solution Generalities.} \label{table:code style 3}
    \end{table}
        
    \noindent\textbf{Summary.} 
    Based on our analysis of coding styles, we draw the following conclusions: 1) Comments on code-based rationales consistently improve performance as they help LLMs understand the logic behind mathematical problems. 2) Descriptive naming conventions should be adopted to enhance the model's understanding of mathematical problems. 3) The hardcoded style of code-based rationales provides a simpler and more effective approach for model learning.

    \subsection{General Coding Instruction is Limited for Math Reasoning}
    The second question is whether coding instructions from general domains, i.e. non-mathematical domains, can improve mathematical ability under PoT reasoning. To investigate this, we train models on three instruction tuning datasets: General Code, Math-Code, and a combination of General Code and Math-Code. 

   \begin{table}[t]
        \centering
        \begin{adjustbox}{width=0.44\textwidth,center}
        \begin{tabular}{l|c|c|c|c|c}
        \hline
         IT Data & Arith & SVAMP & GSM & MATH & Avg. \\
        \hline
        \multicolumn{6}{c}{Zero-Shot} \\
        \hline
        Vanilla Llama & 13.0 & 1.4 & 1.0 & 1.0 & 4.1 \\ 
         + G.C. & \cellcolor{green!50}35.0 & \cellcolor{green!60}36.6 & \cellcolor{green!60}25.3 & \cellcolor{green!50}11.8 & \cellcolor{green!60}27.2 \\ 
         + M.C. & \cellcolor{green!70}80.3 & \cellcolor{green!70}80.9 & \cellcolor{green!70}73.8 & \cellcolor{green!70}32.1 & \cellcolor{green!90}66.8 \\
         + Mix & \cellcolor{green!60}65.3 & \cellcolor{green!70}75.9 & \cellcolor{green!70}71.1 & \cellcolor{green!60}20.3 & \cellcolor{green!80}58.2 \\ 
        \hline
        Vanilla CodeLlama & 19.3 & 7.2 & 1.2 & 2.4 & 7.5 \\ 
        + G.C. & \cellcolor{green!60}70.3 & \cellcolor{green!60}50.9 & \cellcolor{green!60}23.4 & \cellcolor{green!50}14.4 & \cellcolor{green!60}39.8 \\ 
        + M.C. & \cellcolor{green!70}83.7 & \cellcolor{green!70}69.0 & \cellcolor{green!70}58.7 & \cellcolor{green!70}29.2 & \cellcolor{green!80}60.2 \\
        + Mix & \cellcolor{green!70}81.7 & \cellcolor{green!70}70.6 & \cellcolor{green!70}58.1 & \cellcolor{green!60}23.9 & \cellcolor{green!75}58.6 \\
        \hline
        \multicolumn{6}{c}{Few-Shot} \\
        \hline
        Vanilla Llama & 78.0 & 73.1 & 50.0 & 18.0 & 54.8 \\ 
        + G.C. & \cellcolor{red!20}74.7 & \cellcolor{green!10}76.2 & \cellcolor{green!10}51.3 & \cellcolor{green!20}23.0 & \cellcolor{green!05}56.3 \\ 
        + M.C. & \cellcolor{green!20}83.7 & \cellcolor{green!20}80.4 & \cellcolor{green!40}71.6 & \cellcolor{green!40}33.8 & \cellcolor{green!40}67.4 \\
        + Mix & \cellcolor{green!10}78.7 & \cellcolor{green!10}78.0 & \cellcolor{green!20}71.6 & \cellcolor{green!40}34.4 & \cellcolor{green!30}65.7 \\ 
        \hline
        Vanilla CodeLlama & 79.7 & 52.9 & 21.1 & 14.1 & 42.0 \\ 
        + G.C. & \cellcolor{red!10}75.3 & \cellcolor{green!10}54.0 & \cellcolor{green!20}27.3 & \cellcolor{green!20}16.6 & \cellcolor{green!10}43.3 \\ 
        + M.C. & \cellcolor{green!10}80.0 & \cellcolor{green!40}65.8 & \cellcolor{green!60}54.4 & \cellcolor{green!40}25.9 & \cellcolor{green!40}56.5 \\
        + Mix & \cellcolor{green!10}80.7 & \cellcolor{green!35}65.4 & \cellcolor{green!50}51.8 & \cellcolor{green!30}24.7 & \cellcolor{green!30}55.6 \\ 
        \hline
        \end{tabular}
        \end{adjustbox}
        \caption{Performance of coding instruction from different domains. G.C., and M.C. represent General Code, and Math-Code, respectively. Vanilla Llama and Vanilla CodeLlama represent Llama-3.1-8B and CodeLlama-Python-7B, respectively. Cells are green if the instruction tuning boosts the vanilla models' performance, and red if the instruction tuning hurts the performance.}
        \label{table:general code}
    \end{table}

    \begin{figure}[h]
        \centering
        \begin{subfigure}{0.45\textwidth}
            \centering
            \includegraphics[width=\textwidth]{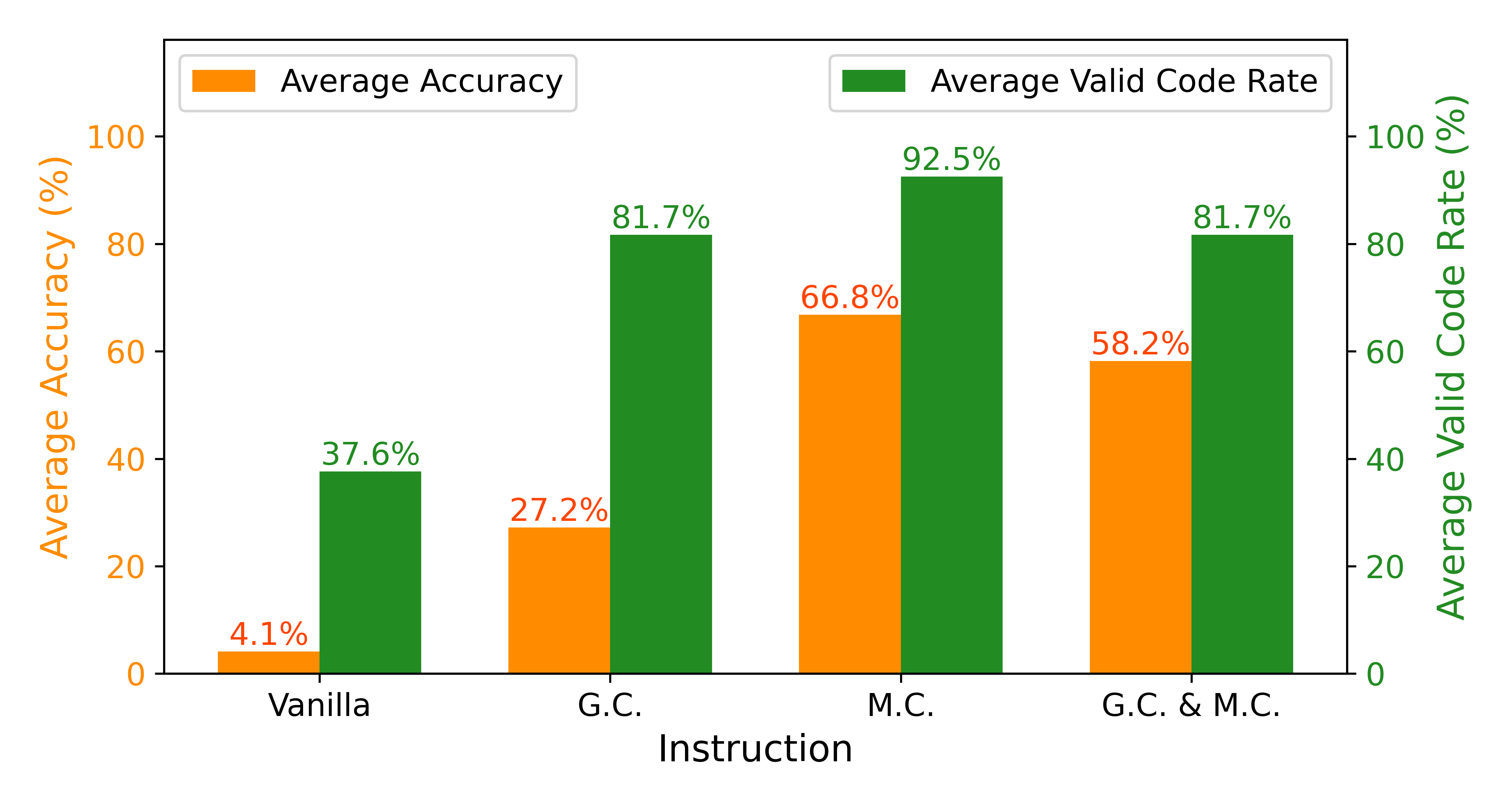}
            \vspace{-15pt}
            \caption{LLama-3.1-8B}
            \label{fig:llama_invalid_code_accuracy}
        \end{subfigure}
        \begin{subfigure}{0.48\textwidth}
            \centering
            \includegraphics[width=\textwidth]{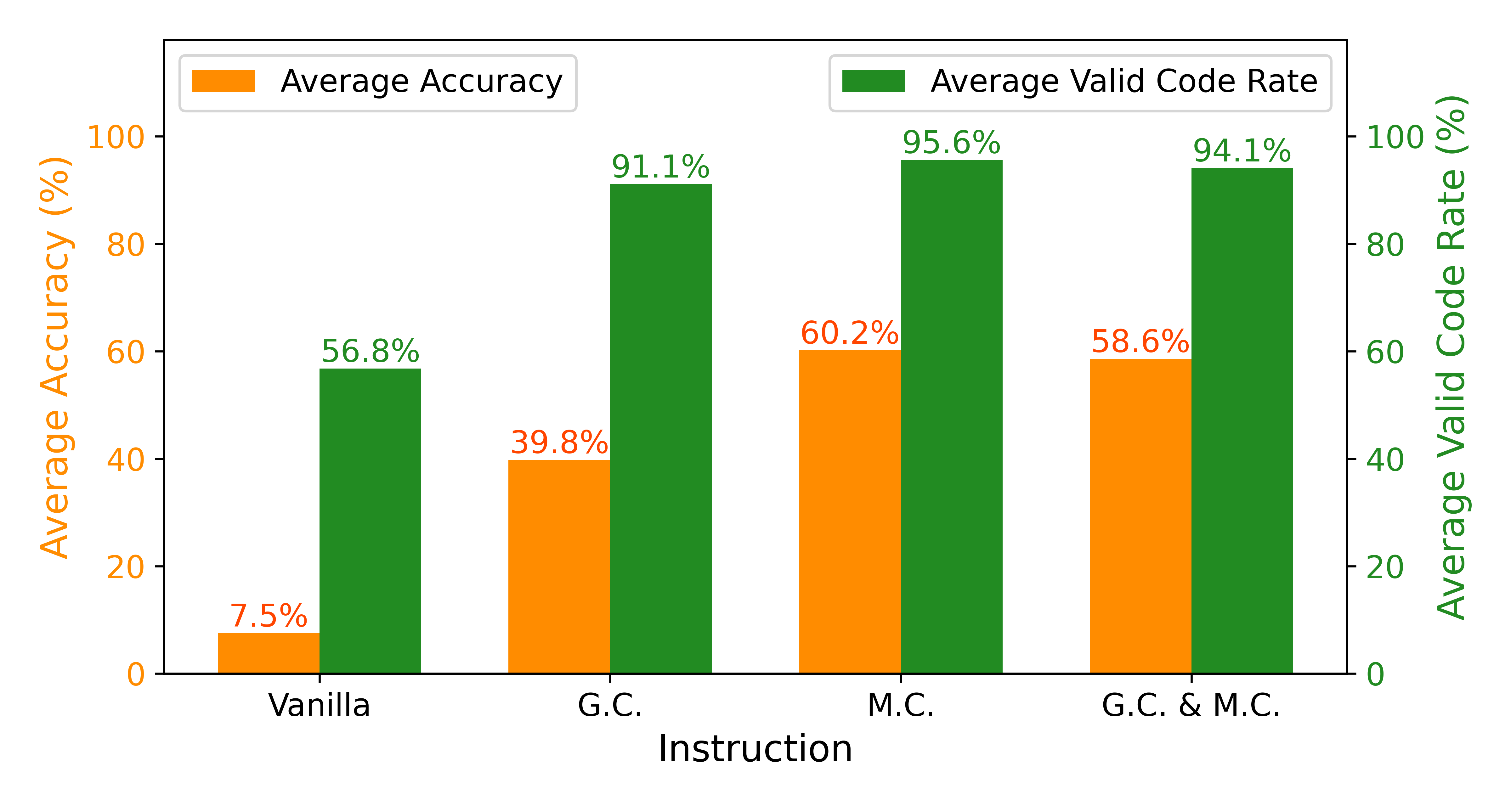}
            \vspace{-15pt}
            \caption{CodeLlama-Python-7B}
            \label{fig:codellama_invalid_code_accuracy}
        \end{subfigure}
        \caption{Average accuracy and average valid code rate across the evaluation datasets under zero-shot.}
        \label{fig:valid_code_rate_vs_accuracy}
    \end{figure}

    Prior studies \cite{wang2023far, zhang2024unveiling} indicate that, under CoT reasoning, instruction tuning on general code, i.e., non-math-specific data can have trivial or even negative effects on mathematical reasoning abilities. Under PoT reasoning, our experiments reveal a different trend, which is presented in Table \ref{table:general code}. 
    
    In both zero-shot and few-shot settings, training solely on General Code mostly enhances the models' performance in solving math problems. This improvement is primarily attributed to the models' enhanced ability to generate valid code and follow instructions, making the models better suited for PoT reasoning compared to the base model. We define the valid code rate as the percentage of executable code generated by a model. Figure~\ref{fig:valid_code_rate_vs_accuracy} presents the accuracy and valid code rate under zero-shot settings. The results demonstrate that training on General Code significantly improves the models' ability to generate valid code and adhere to instructions, leading to higher accuracy in solving mathematical questions.

    However, models solely trained on Math-Code achieve the best performance, and incorporating General Code alongside Math-Code reduces its effectiveness. This implies that non-task-relevant coding instructions, i.e., General Code, distract the models and hinder their mathematical reasoning capabilities. These findings suggest that coding instructions from general domains offer limited benefits for enhancing mathematical reasoning. 
    
    \subsection{Math-Text is Not Always a Supplement for Math-Code}

    Recent works \cite{yue2023mammoth, wang2023mathcoder} commonly employ hybrid training on both textual and code-based rationales to enhance mathematical reasoning. According to \citet{yue2023mammoth}, textual rationales contribute to general language-based reasoning, particularly for scenarios where PoT reasoning struggles, such as abstract reasoning in multiple-choice questions.

    In this study, we delve deeper by focusing on math questions that require concrete calculations, excluding abstract reasoning, and investigate whether Math-Text consistently serves as a supplement to Math-Code. To ensure a fair comparison, we trained models on three instruction tuning datasets: Math-Text, Math-Code, which contain identical mathematical questions, differing only in the rationales provided, and a combination of both.

    The results presented in Table \ref{table:MT-MC} show different trends between general-purpose and code-specialized models. For the general-purpose model, Llama-3.1-8B, incorporating Math-Text alongside Math-Code slightly improves performance by enhancing the model’s ability to comprehend mathematical problems in a language-based manner. However, for code-specialized models, this addition negatively impacts performance, as these models are inherently optimized for code generation. The inclusion of textual rationales may interfere with their specialized capabilities. This finding suggests that when mathematical questions include code-based rationales, augmenting them with textual explanations is beneficial only for models that are proficient in textual reasoning.

    \begin{table}[t]
        \centering
        \begin{adjustbox}{width=0.45\textwidth,center}
        \begin{tabular}{c|c|cccc|c}
        \toprule
         Model & IT Data & Arith & SVAMP & GSM & MATH & Avg \\
        \hline
        \multicolumn{7}{c}{Zero-Shot} \\
        \hline
         \multirow{3}{*}{\shortstack{Llama-3.1-\\8B}} & M.T. & 38.7 & 62.2 & 54.6 & 20.6 & 44.0 \\ 
         & M.C. & 80.3 & 80.9 & 73.8 & 32.1 & 66.8 \\
         & Mix & 80.3 & 81.1 & 76.0 & 34.4 & \textbf{67.9} \\ 
        \hline
        \multirow{3}{*}{\shortstack{CodeLlama-\\7B}} & M.T. & 22.7 & 44.8 & 32.4 & 7.9 & 26.9 \\ 
        & M.C. & 83.7 & 69.0 & 58.7 & 29.2 & \textbf{60.2} \\
        & Mix & 79.3 & 67.8 & 59.1 & 28.1 & 58.6 \\
        \hline
        \multicolumn{7}{c}{Few-Shot} \\
        \hline
        \multirow{3}{*}{\shortstack{Llama-3.1-\\8B}} & M.T. & 48.3 & 67.5 & 62.7 & 21.8 & 50.1 \\ 
        & M.C. & 83.7 & 80.4 & 71.6 & 33.8 & 67.4 \\
        & Mix & 83.0 & 79.8 & 73.0 & 35.4 & \textbf{67.8} \\ 
        \hline
        \multirow{3}{*}{\shortstack{CodeLlama-\\7B}} & M.T. & 31.0 & 48.4 & 26.7 & 7.9 & 28.5 \\ 
        & M.C. & 80.0 & 65.8 & 54.4 & 25.9 & \textbf{56.5} \\
        & Mix & 78.3 & 65.1 & 53.4 & 24.5 & 55.3 \\ 
        \bottomrule
        \end{tabular}
        \end{adjustbox}
        \caption{Performance of coding instruction from textual and code-based rationales. M.T, and M.C. represent Math-Code, and Math-Code, respectively. M.T. uses CoT prompting inference while others use PoT prompting inference.}
        \label{table:MT-MC}
    \end{table}

    \begin{figure*}[t]
        \centering
        \includegraphics[width=0.98\textwidth]{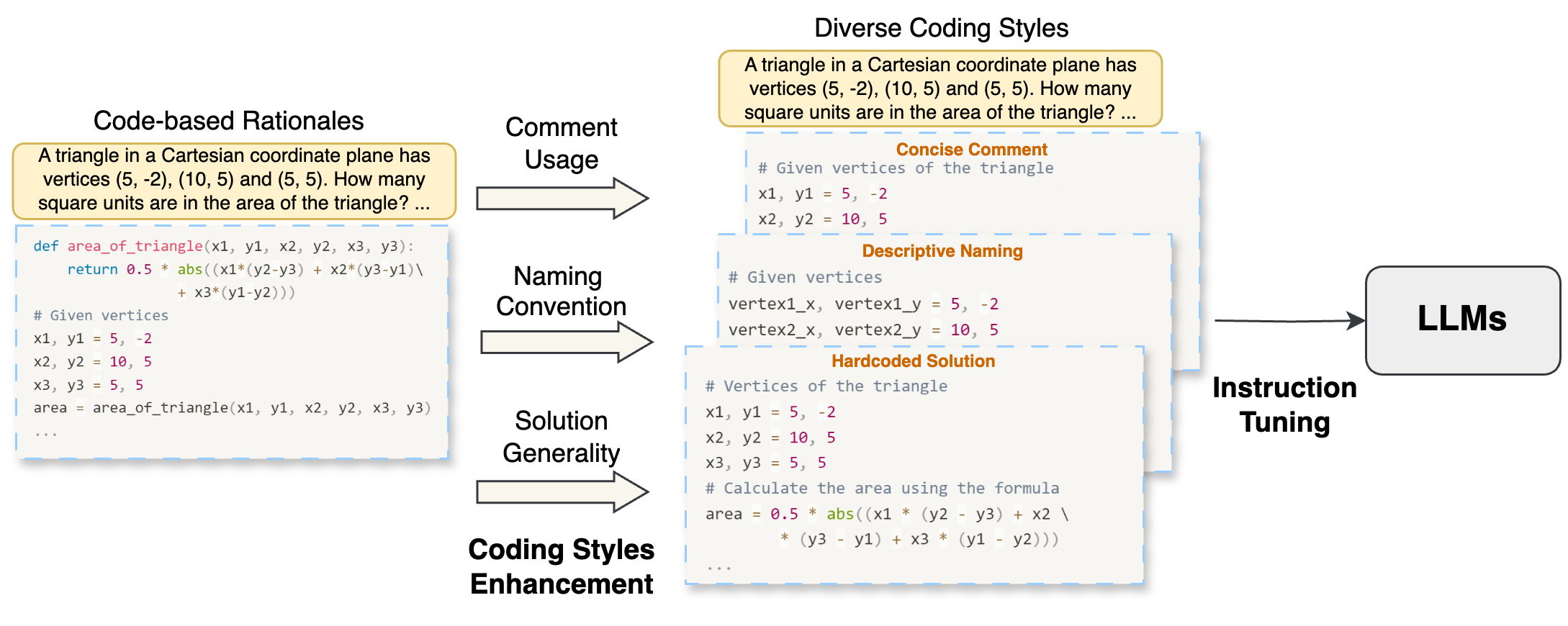}
        \caption{Overview of CoinMath framework. CoinMath generates three distinct variations of code-based rationales with advantageous coding attributes—Concise Comment, Descriptive Naming, and Hardcoded Solution—and ensembles them for LLM instruction tuning.}
        \label{fig:coinmath}
    \end{figure*}

\section{CoinMath}

    Building on the insights from our systematic study on coding instruction, we introduce CoinMath, a strategy that enhances coding instructions by incorporating beneficial coding styles to improve the mathematical reasoning capabilities of LLMs.

    \subsection{Method}
    The CoinMath framework is depicted in Figure~\ref{fig:coinmath}. As discussed in the previous section, LLMs demonstrate varying performance depending on the coding styles of mathematical rationales. Among the evaluated coding styles, Concise Comment performs best in the Comment Usage category, Descriptive Naming outperforms in Naming Conventions, and Hardcoded Solution outperforms in Solution Generalities. However, while these styles outperform their counterparts within their respective attributes, no single coding style proves universally optimal for all mathematical questions. For instance, as shown in Section~\ref{subsec: study on coding styles}, Hardcoded Solution, while better than Generalized Solution in the Solution Generality attribute, performs poorly on the SVAMP dataset for Llama-3.1-8B. This suggests that a fixed coding style may not be ideal for all mathematical problems.

    Instead of creating only one code-based rationale for a math question with fixed coding attributes, CoinMath generates three variations of rationales with diverse beneficial coding styles, using the same strategy outlined in Section~\ref{subsec: study on coding styles}. Each variation emphasizes one specific coding style attribute: concise comments, descriptive naming conventions, or hardcoded solutions. CoinMath then ensembles these three variations of rationales with beneficial coding attributes to maximize the enhancement of models' mathematical reasoning.
    
    Additionally, CoinMath excludes coding instructions from general domains and mathematical textual rationales, as their contributions to improving mathematical reasoning are minimal or even negative.

    \begin{table*}[h]
        \centering
        \begin{adjustbox}{width=0.97\textwidth,center}
        \begin{tabular}{l|l|c|c|c|c|c|c}
        \toprule
        Model & Base & Prompt & Arithmetic & SVAMP & GSM & MATH & Avg \\
        \midrule
        \multicolumn{7}{l}{\emph{Non-Math-Specific Model (<8B)}} \\
        \midrule
        Llama-3.1 & - & CoT & 36.7 & 45.7 & 17.9 & 8.6 & 27.3\\
        Llama-3.1-Instruct & - & CoT & 51.7 & 81.1 & 75.1 & 45.7 & 63.4\\
        CodeLlama-Python & - & Hybrid  & 24.3 & 21.9 & 3.7 & 4.9 & 13.7 \\ 
        \midrule
        \multicolumn{7}{l}{\emph{Math-specific Model (<8B)}} \\
        \midrule
        Qwen2-Math & Qwen2 & CoT & 76.3 & 91.4 & 78.8 & 71.6 & 79.5\\
        Qwen2-Math-Instruct & Qwen2 & CoT & 65.0 & 87.0 & 79.2 & 70.2 & 75.4 \\
        WizardMath-V1.1 & Llama-2 & CoT & 38.3 & 72.5 & 67.4 & 33.5 & 52.9 \\
        MathCoder-L$^{\dagger}$ & Llama-2 & - & - & 71.5 & 64.2 & 23.3 & - \\
        MathCoder-CL$^{\dagger}$ & CodeLlama & - & - & 70.7 & 67.8 & 30.2 & - \\
        MAmmoTH & Llama-2 & Hybrid & 13.0 & 65.3 & 51.9 & 31.5 & 40.4 \\ 
        MAmmoTH-Coder & CodeLlama & Hybrid & 31.0 & 54.2 & 31.8 & 20.5 & 34.4 \\
        \hline
        MAmmoTH & Llama-3.1 & Hybrid & 71.4 & 81.3 & 73.0 & 39.1 & 66.2 \\
        MAmmoTH-Coder & CodeLlama-Python & Hybrid & 72.7 & 67.0 & 52.9 & 21.8 & 53.6 \\
        \hline
        CoinMath & Llama-3.1 & Hybrid & 77.0 (\textcolor{red}{+5.6}) & 83.1 (\textcolor{red}{+1.8}) & 76.4 (\textcolor{red}{+3.4}) & 40.3 (\textcolor{red}{+1.2}) & 69.2 (\textcolor{red}{+3.0}) \\
        CoinMath & CodeLlama-Python & Hybrid & 80.7 (\textcolor{red}{+8.0}) & 75.1 (\textcolor{red}{+8.1}) & 62.0 (\textcolor{red}{+9.1}) & 31.8 (\textcolor{red}{+10.0}) & 62.4 (\textcolor{red}{+8.8})\\
        \bottomrule
        \end{tabular}
        \end{adjustbox}
        \caption{Zero-shot performance on mathematical evaluation datasets. Red numbers highlight the improvement compared with the same base models trained on MathInstruct, i.e., MAmmoTH models. $^{\dagger}$ means the results are from the corresponding papers.} \label{table:final result}
    \end{table*}
    
    \subsection{Experimental Setup}
    We scale up the base instruction tuning dataset using the complete set of code-based rationales from MathInstruct~\cite{yue2023mammoth}, which contains 73k math questions with code-based rationales. Subsequently, we leverage this scaled dataset to generate code-based rationales with diverse coding styles using GPT-4o.
    We use the same evaluation datasets as in Section~\ref{sec:coding instruction study}.
    
    For comparison, we select representative models from the following three categories:
    \textbf{Base models}: We consider LLama-3.1 \cite{dubey2024llama} and CodeLlama-Python-7B \cite{roziere2023code} as our base models. 
    \textbf{Instruct models}: We include CodeLlama-3.1-8B-Instruct \cite{dubey2024llama} as a representative instruct-tuned model. 
    \textbf{Math-specific models}: These models are fine-tuned specifically for solving math problems, including WizardMath \cite{luo2023wizardmath}, MathCoder \cite{wang2023mathcoder}, and MAmmoTH \cite{yue2023mammoth}. Since our instruction-tuning dataset with diverse coding styles is derived from MathInstruct which MAmmoTH utilizes, we further fine-tuned LLama-3.1-8B and CodeLlama-Python-7B on MathInstruct to ensure consistency in base models used for comparison. This approach establishes a strong baseline for assessing the improvements introduced by CoinMath. Additionally, we include Qwen2-Math-7B and Qwen2-Math-7B-Instruct \cite{yang2024qwen2}, which are heavily optimized for mathematics and demonstrate superior performance, rivaling even closed-source models (e.g., GPT-4).

    We evaluate the models' zero-shot performance and select the best result from two prompting approaches: CoT prompting and hybrid prompting. The hybrid prompting approach first uses PoT prompting and resorts to CoT prompting only if PoT fails to generate a result.

    \subsection{Results and Analysis}
    The experimental results are presented in Table~\ref{table:final result}. CoinMath significantly outperforms MAmmoTH after learning from code-based rationales that incorporate concise comments, descriptive naming conventions, and hardcoded solutions. 
    Additionally, both CoinMath and Llama-3.1-Instruct are built on the Llama-3.1 base model and are instruct-tuned for mathematical reasoning, with Llama-3.1-Instruct further fine-tuned on other instruction-tuning topics. CoinMath surpasses Llama-3.1-Instruct except for the MATH dataset, demonstrating the effectiveness of our approach.

    It is worth noting that the performance on the Arithmetic dataset, an out-of-domain evaluation dataset, decreases compared to the results in Section~\ref{sec:coding instruction study} as the training set is scaled up. In addition, it is reasonable that our models fall behind the Qwen2-Math models, as those models undergo extensive training for mathematics across pre-training, instruction tuning, and reinforcement learning stages while CoinMath focuses solely on the instruction tuning stage.

\begin{figure}[t]
    \centering
    \includegraphics[width=0.47\textwidth]{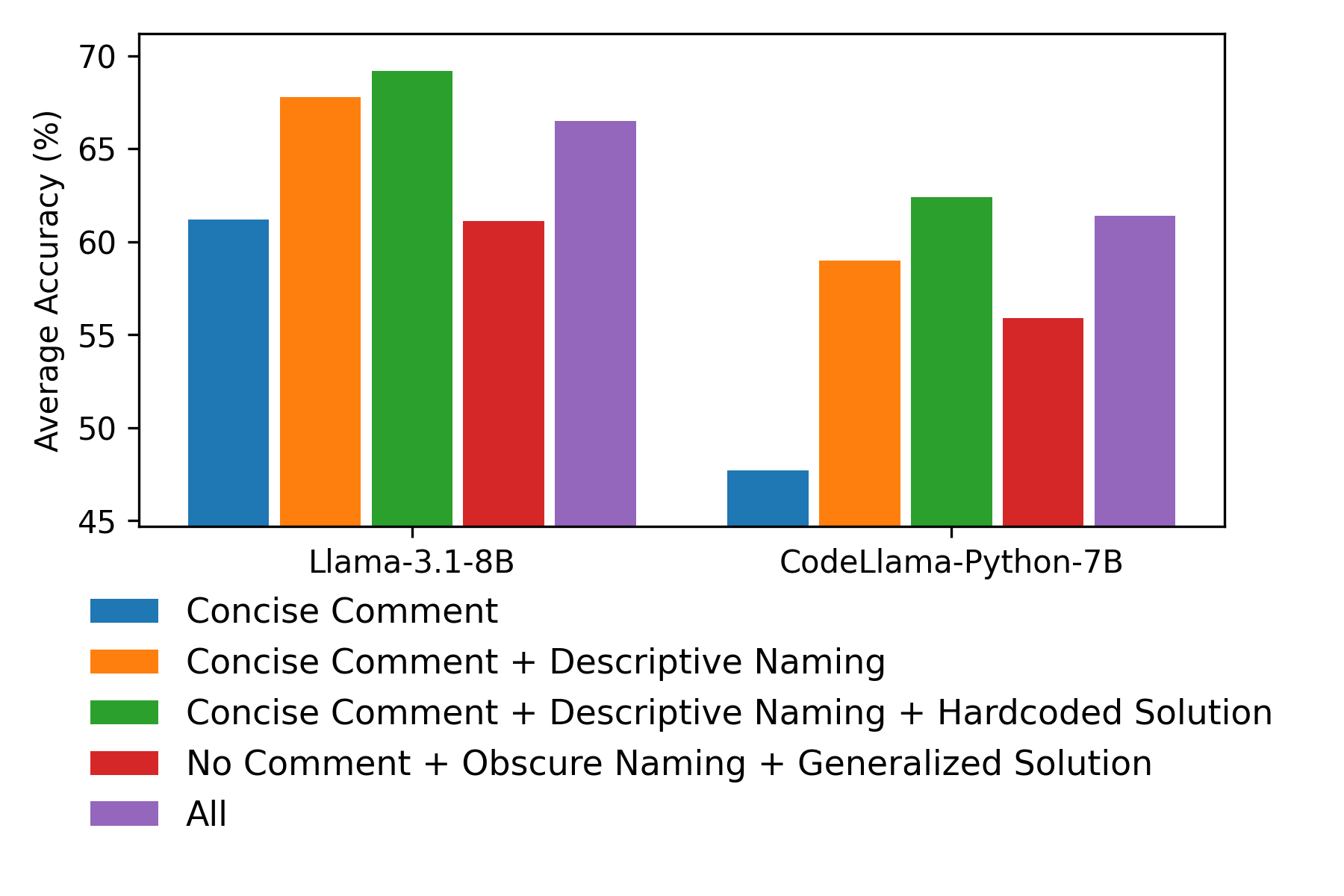}
    \vspace{-12pt}
    \caption{Average accuracy of the models using code-based rationales with different combinations of styles. The average accuracy is calculated across the Arithmetic, SVAMP, GSM, and MATH datasets.}
    \label{fig:ablation_study}
    \end{figure}

    \noindent\textbf{Ablation Study.}
    We investigate the impact of combining different coding styles on model performance. Figure~\ref{fig:ablation_study} presents the average accuracy across our evaluation datasets for various combinations of coding styles. Detailed performance metrics for each individual evaluation dataset are provided in Appendix~\ref{sec:appendix_ablation_study}. Our results show that incorporating concise comments, descriptive naming conventions, and hardcoded solutions consistently enhances the model's performance in solving math problems. In contrast, their counterparts—no comment, obscure naming conventions, and generalized solutions—yield relatively inferior performance, achieving results similar to training solely with concise comments. Finally, combining all coding styles does not achieve performance as high as the combination of concise comments, descriptive naming conventions, and hardcoded solutions, further proving that these three attributes are particularly effective for code-based rationales in improving the model's mathematical reasoning ability.
    
\section{Conclusion}

    We conduct a systematic study to investigate how coding instructions can effectively enhance the mathematical reasoning abilities of LLMs. Based on these findings, we propose CoinMath, which focuses specifically on mathematical code-based rationales and combines mathematical coding instructions with concise comments, descriptive naming, and hardcoded solutions to boost the mathematical reasoning capabilities of LLMs. Experimental results demonstrate that CoinMath outperforms the baseline model, MAmmoTH, one of the SOTA models, by an average improvement of 5.9\% in accuracy.

\section*{Limitations}
    We focus on evaluating LLMs' mathematical reasoning abilities on datasets requiring concrete calculations, including Arithmetic, SVAMP, GSM, and MATH. However, for other types of math questions involving abstract reasoning or mathematical knowledge, such as abstract algebra, and theorem comprehension. code-based solutions may be less effective. This underscores the importance of extending our work to these tasks to gain insights across a broader range of mathematical question types.

\bibliography{anthology,custom}
\bibliographystyle{acl_natbib}

\appendix
\section{Statistics of Instruction Tuning Datasets}

Table~\ref{table:it data} presents the statistics of the three base instruction-tuning datasets used in our study on coding instruction. Math-Code and Math-Text share the same set of mathematical questions but differ in their rationales, with Math-Code providing code-based rationales and Math-Text offering textual ones.

    \begin{table}[h]
        \centering
        \begin{adjustbox}{width=0.5\textwidth,center}
        \begin{tabular}{c|cccc}
        \toprule
        \textbf{IT Dataset} & \textbf{\# Sample} & \textbf{Characteristics} & \textbf{Annotation} \\
        \midrule
        \multirow{2}{*}{Math-Code} & \multirow{2}{*}{26k} & \multirow{2}{*}{\shortstack{Math questions with \\ code-based rationales}} & \multirow{2}{*}{GPT4} \\
        \\
        \multirow{2}{*}{Math-Text} & \multirow{2}{*}{26k} & \multirow{2}{*}{\shortstack{Math questions with \\ texutal rationales}} & \multirow{2}{*}{Human \& Llama} \\
        \\
        \multirow{2}{*}{General Code} & \multirow{2}{*}{21k} & \multirow{2}{*}{\shortstack{Coding instruction \\for general tasks}} & \multirow{2}{*}{Llama} \\
        \\
        \bottomrule
        \end{tabular}
        \end{adjustbox}
        \caption{Statistics of Math-Text, Math-Code and General Code} \label{table:it data}
    \end{table}

\section{Sample Questions from Evaluation Datasets}
\label{sec:appendix_eva_examples}

\begin{figure}[h]
    \centering
    \begin{subfigure}{0.48\textwidth}
        \centering
        \includegraphics[width=\textwidth]{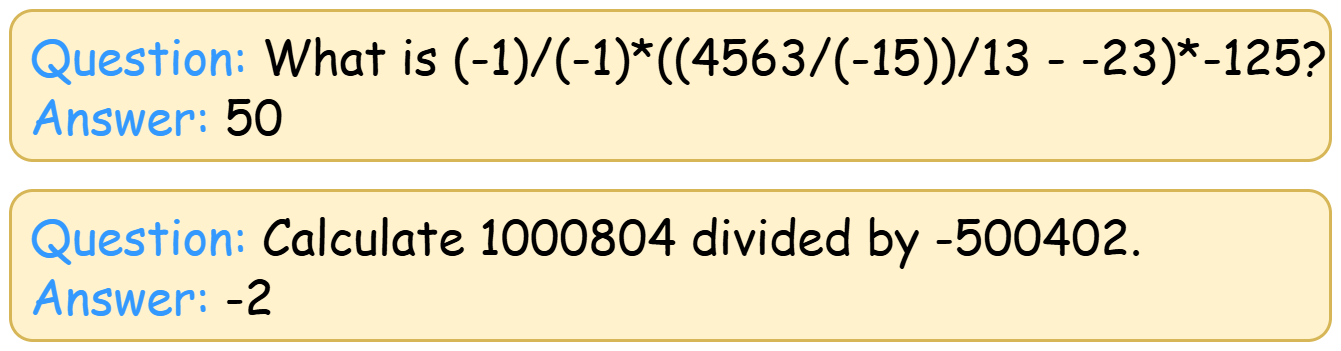}
        \caption{Arithmetic}
    \end{subfigure}
    \hfill
    \begin{subfigure}{0.48\textwidth}
        \centering
        \includegraphics[width=\textwidth]{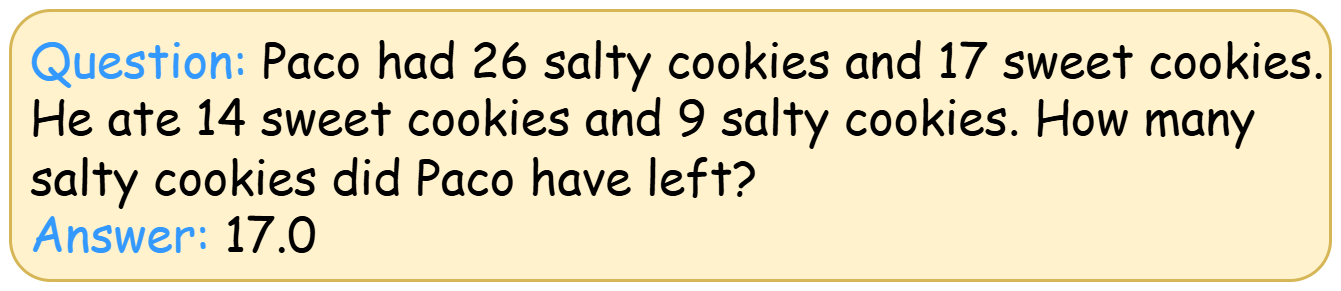}
        \caption{SVAMP}
    \end{subfigure}
    \hfill
    \begin{subfigure}{0.48\textwidth}
        \centering
        \includegraphics[width=\textwidth]{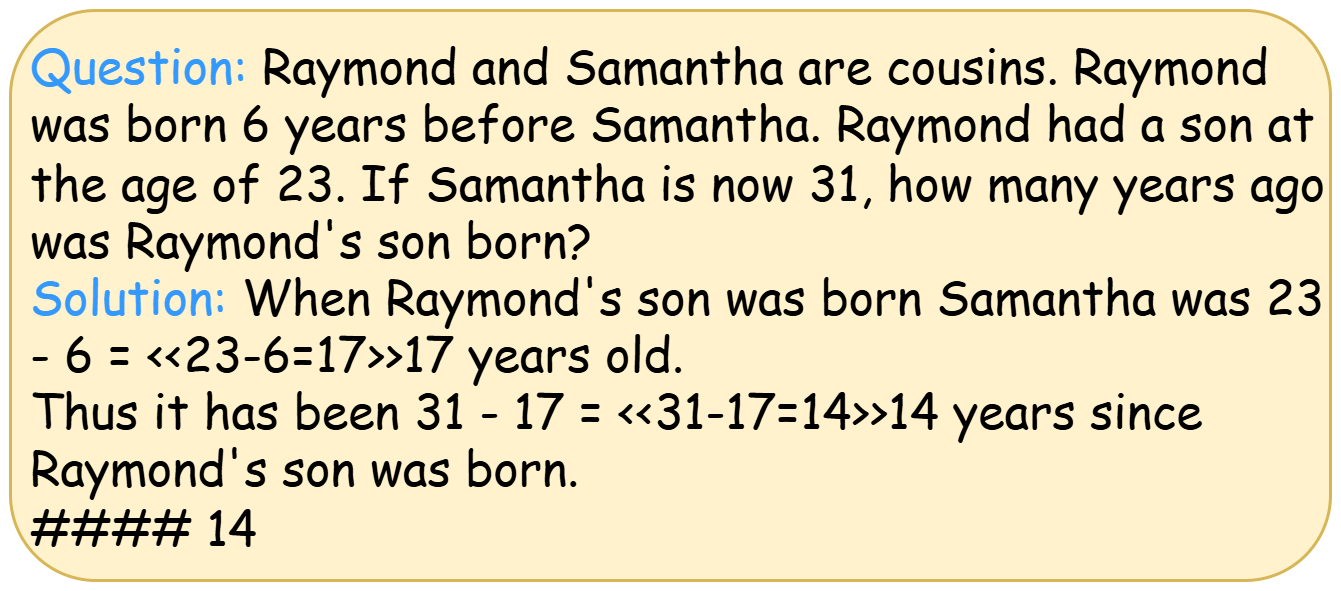}
        \caption{GSM}
    \end{subfigure}
    \hfill
    \begin{subfigure}{0.48\textwidth}
        \centering
        \includegraphics[width=\textwidth]{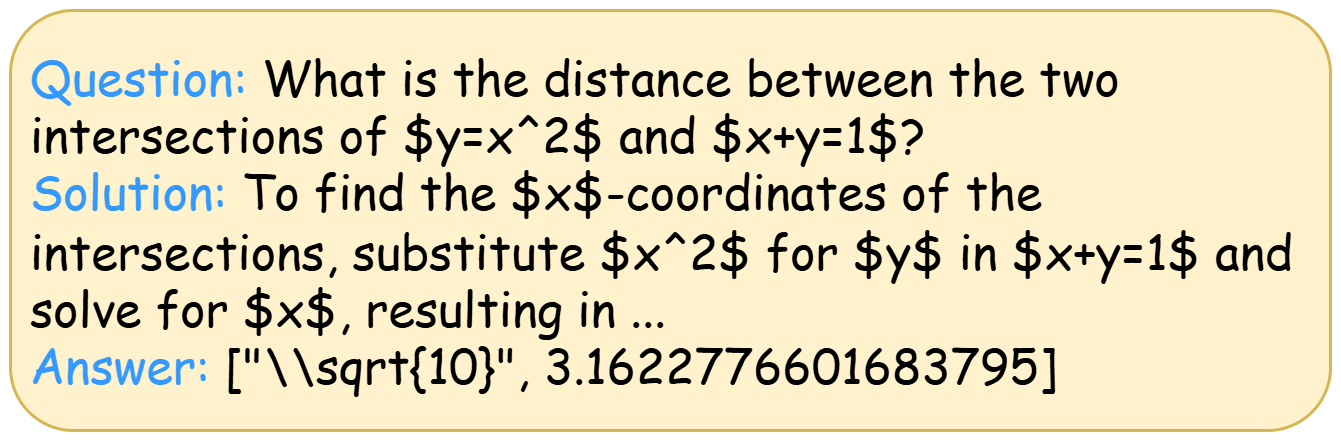}
        \caption{MATH}
    \end{subfigure}
    \caption{Sample questions from the evaluation datasets}
    \label{fig:example_eva_set}
\end{figure}

Figure~\ref{fig:example_eva_set} shows the samples from Arithmetic, SVAMP, GSM and MATH.

\begin{figure}[h]
    \centering
    \begin{subfigure}{0.5\textwidth}
        \centering
        \includegraphics[width=\textwidth]{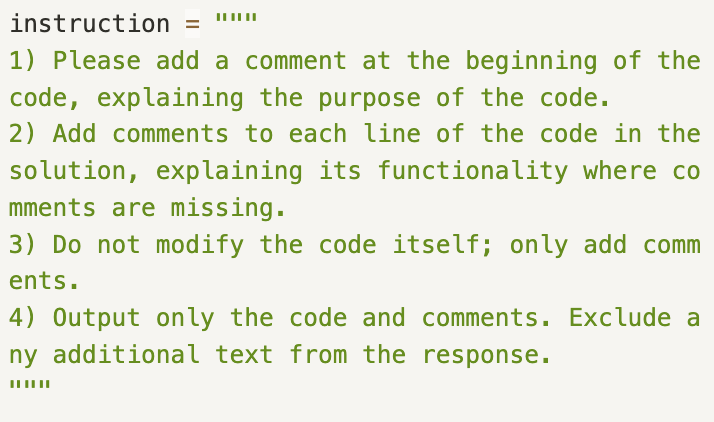}
        \caption{Instruction for generating detailed comments}
    \end{subfigure}
    \hfill
    \begin{subfigure}{0.5\textwidth}
        \centering
        \includegraphics[width=\textwidth]{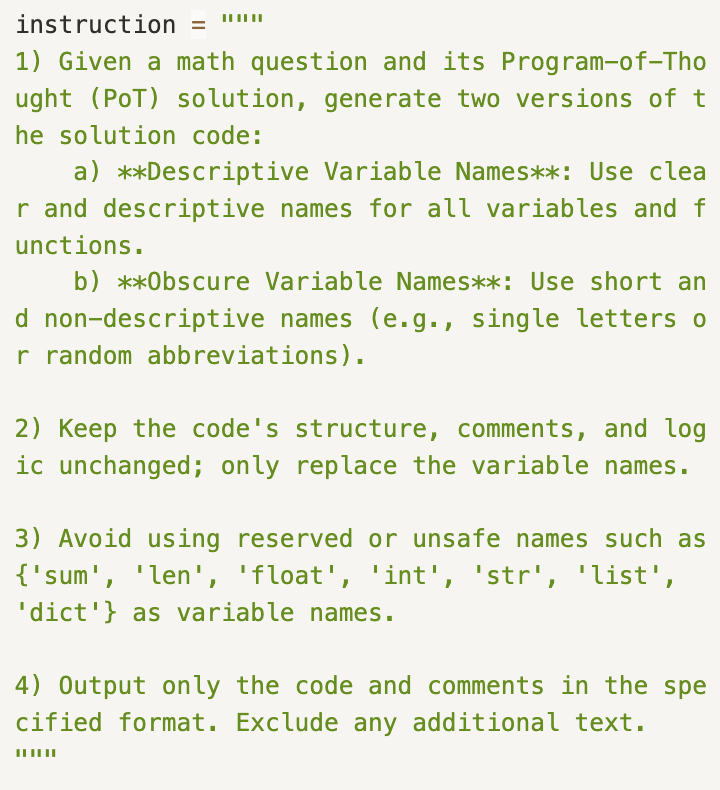}
        \caption{Instruction for applying different naming conventions}
    \end{subfigure}
    \hfill
    \begin{subfigure}{0.5\textwidth}
        \centering
        \includegraphics[width=\textwidth]{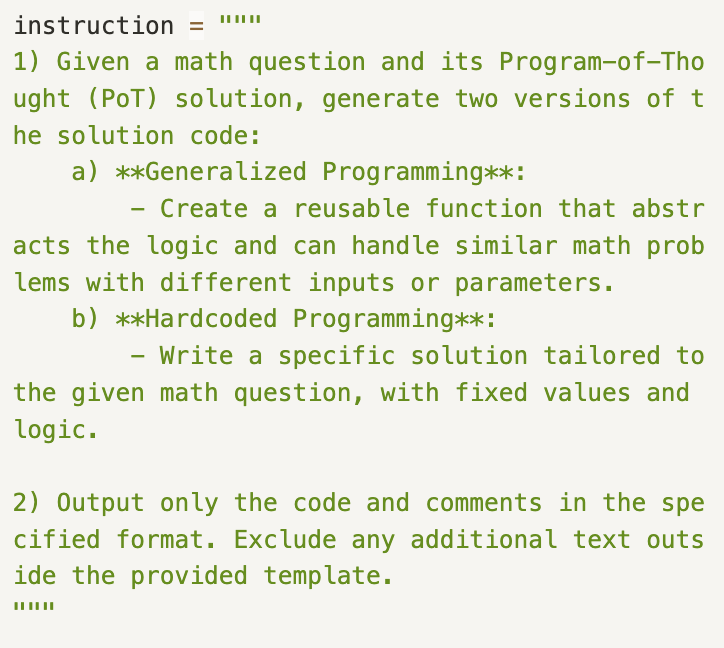}
        \caption{Instruction for generating generalized and hardcoded solutions}
    \end{subfigure}
    \caption{Instruction for generation diverse coding style rationales}
    \label{fig:appendix_instruction}
\end{figure}

\section{Prompts for Generating Diverse Coding Style Rationales}
\label{sec:appendix_prompt}
Figure~\ref{fig:appendix_instruction} shows the instruction we used for generating code-based rationales with different coding styles.

\section{Instruction Tuning Details}
\label{sec:appendix_instruction_tuning_details}
We perform instruction tuning of the LLMs following the implementation of TULU~\cite{wang2023far, ivison2023camels}. The models are fine-tuned using LoRA~\cite{hu2021lora} with a rank of 64 and a total batch size of 128.

\section{Results of Mixing Coding Styles}
\label{sec:appendix_ablation_study}
Table~\ref{table:ablation study} demonstrates the performance of different combinations of coding styles for each evaluation dataset.

\begin{table*}[htb]
    \centering
    \scalebox{0.8}{
    \begin{tabular}{l|l|cccc|c}
    \toprule
    Model & IT Data & Arithmetic & SVAMP & GSM & MATH & Average \\
    \hline
    \multirow{5}{*}{\shortstack{Llama-\\3.1-8B}} & Concise Comment & 58.0 & 75.5 & 72.8 & 38.6 & 61.2 \\
    & Concise Comment + Descriptive Naming & 81.7 & 80.5 & 71.3 & 37.6 & 67.8 \\
    & Concise Comment + Descriptive Naming + Hardcoded Solution & 77.0 & 83.1 & 76.4 & 40.3 & 69.2 \\
    \cline{2-7}
    & No Comment + Obscure Naming + General Solution & 48.3 & 81.4 & 75.2 & 39.4 & 61.1 \\
    & All & 72.3 & 83.0 & 75.0 & 35.8 & 66.5 \\
    \hline
    \multirow{5}{*}{\shortstack{CodeLlama-\\7B}} & Concise Comment & 44.3 & 68.6 & 56.6 & 21.4 & 47.7 \\
    & Concise Comment + Descriptive Naming & 80.3 & 69.6 & 56.6 & 29.5 & 59.0 \\
    & Concise Comment + Descriptive Naming + Hardcoded Solution & 80.7 & 75.1 & 62.0 & 31.8 & 62.4 \\
    \cline{2-7}
    & No Comment + Obscure Naming + General Solution & 66.0 & 69.9 & 60.2 & 27.4 & 55.9 \\
    & All & 79.3 & 73.7 & 60.7 & 31.7 & 61.4 \\
    \bottomrule
    \end{tabular}}
    \caption{Ablation Study of Coding Styles. Evaluation is under zero-shot setting} \label{table:ablation study}
\end{table*}

\end{document}